\documentclass[conference]{IEEEtran}
\usepackage{cite}
\usepackage{graphicx}
\usepackage{amsmath}
\usepackage{amssymb}
\usepackage{algorithmic}
\usepackage{algorithm}
\usepackage[caption=false,font=footnotesize]{subfig}
\usepackage{stfloats}
\usepackage{url}
\usepackage{tikz}
\usetikzlibrary{bayesnet}
\usepackage{float}
\usepackage{csvsimple}
\floatname{algorithm}{Algorithm}

\begin{document}
\title{Cadre Modeling: Simultaneously Discovering Subpopulations and Predictive Models}

\author{\IEEEauthorblockN{Alexander New\textsuperscript{1,2}, Curt Breneman\textsuperscript{3}, Kristin P. Bennett\textsuperscript{1,2}}
\IEEEauthorblockA{
\textsuperscript{1}Institute for Data Exploration and Application\\
\textsuperscript{2}Department of Mathematical Sciences\\
\textsuperscript{3}Department of Chemistry and Chemical Biology\\
Rensselaer Polytechnic Institute\\
Troy, NY\\
\{newa, brenec, bennek\}@rpi.edu}}
%\and
%\IEEEauthorblockN{Curt Breneman}
%\IEEEauthorblockA{Department of Chemistry and Chemical Biology\\
%Rensselaer Polytechnic Institute\\
%Troy, NY\\
%brenec@rpi.edu}}
%\and 
%\author{\IEEEauthorblockN{Kristin P. Bennett}
%\IEEEauthorblockA{Institute for Data Exploration and Application\\
%Rensselaer Polytechnic Institute\\
%Troy, NY\\
%bennek@rpi.edu}

\maketitle

\begin{abstract}
We consider the problem in regression analysis of identifying subpopulations that exhibit different patterns of response, where each subpopulation requires a different underlying model. Unlike statistical cohorts, these subpopulations are not known \emph{a priori}; thus, we refer to them as cadres. When the cadres and their associated models are interpretable, modeling leads to insights about the subpopulations and their associations with the regression target. We introduce a discriminative model that simultaneously learns cadre assignment and target-prediction rules. Sparsity-inducing priors are placed on the model parameters, under which independent feature selection is performed for both the cadre assignment and target-prediction processes. We learn models using adaptive step size stochastic gradient descent, and we assess cadre quality with bootstrapped sample analysis. We present simulated results showing that, when the true clustering rule does not depend on the entire set of features, our method significantly outperforms methods that learn subpopulation-discovery and target-prediction rules separately. In a materials-by-design case study, our model provides state-of-the-art prediction of polymer glass transition temperature. Importantly, the method identifies cadres of polymers that respond differently to structural perturbations, thus providing design insight for targeting or avoiding specific transition temperature ranges. It identifies chemically meaningful cadres, each with interpretable models. Further experimental results show that cadre methods have generalization that is competitive with linear and nonlinear regression models and can identify robust subpopulations.
\end{abstract}

%Our novel model softly partitions the observations into cadres defined using only a few features. Within these cadres, the behavior of the target variable is more simply modeled than it is on the population as a whole.

%\textbf{ASN Note:} The abstract is 1622 characters. The IJCNN limit is 1750.

%\textbf{ASN Note:} Subpopulation instead of group or partition

%\textbf{ASN Note:} Cadre size vs. generalization error for cheminformatics

%\textbf{ASN Note:} High-level: present as new idea (cadres), not as incremental idea (sparse probabilistic regression cadres)

%\textbf{ASN Note:} Repeat that regression is special case

\IEEEpeerreviewmaketitle

%\textbf{ASN note:} Once we finalize what the introduction should say, we should probably find more references to put in it.

%\textbf{ASN note:} We could include references in the related works to the epidemiology stuff where analysts know to assign different models to different cohorts as a contrast to the type of problem we are considering.

%\textbf{ASN note:} If we are not pushing interpretability as much anymore, we can drop the $\tau$ statistic -- it was always kind of made up.

%\textbf{ASN note:} Make sure location of citation when at end of sentence is consistent.

%\textbf{ASN note:} Sometimes we say ``response'', and sometimes we say ``target''. Is that okay, or should we try to stick to one?

\section{Introduction}
In many real-world analytics tasks, observations are divided into different cohorts using simple criteria, and a different predictive model is created for each cohort. For example, in insurance, it is well known that young adults have different accident rates than do adults. Thus, one would make different predictive models of accident rates for different age subpopulations\cite{life_insurance}. Similarly, an epidemiologist may make different models for survey subjects of different ethnicities or socioeconomic statuses\cite{epidemiology1}. 

For most problems, however, appropriate subpopulations are not known \emph{a priori}. Identifying appropriate subpopulations becomes a valuable aspect of the learning task. Consider the task in drug discovery of predicting the toxicity of a molecule. Different types of molecules have different mechanisms of toxicity, so understanding these classes of molecules and the different factors that influence their toxicity can help accelerate the drug discovery process\cite{drugs}. In these examples and others, only a few features are used in defining the subpopulations. Thus, a method for discovering new subpopulations must be able to perform feature selection.

If the possible existence of these subpopulations is ignored, a more complicated model such as a kernel support vector machine (SVM)\cite{svms} may be required to attain a good fit. However, it is far more informative when this single complicated model can be replaced with a set of simpler ones\cite{ensemble}\cite{jordanHME}, each associated with a subpopulation of the dataset. The proposed learning problem is to jointly discover these subpopulations and estimate their prediction models. To maximize the insight the model provides, we seek to make interpretable models for both characterizing the subpopulations and predicting the target within each subpopulation.

%\textbf{ASN note:} Can we choose a more informative word than ``value'' here?

%We seek to develop machine learning models for regression problems that are more interpretable than nonlinear techniques such as kernel support vector machines\cite{svms} or deep neural networks\cite{deeplearning} while being more powerful than standard linear models. For these methods, model predictions are meant to augment, rather than replace, human decisions and insight. In this case, the learning goal goes beyond generalization: the domain expert should be able to understand how a machine learning model makes its predictions. The proposed methods are also effective on problems when there is insufficient data for complicated nonlinear models to perform well. 

Our way of solving this class of problem is the supervised cadre model (SCM). An SCM divides the input space into probabilistic subpopulations that we call cadres. Cadres are similar to clusters\cite{lloyd}, but they are able to perform feature selection and weighting. They are also selected so as to be easily assigned a predictive model.

An SCM has two major components. The first is the cadre membership function, which assigns every point in input space a distribution characterizing that point's probability of belonging to each cadre. The second component of an SCM is a set of target-prediction models. In this work, we focus on the case where the target-prediction models are linear. Crucially, the cadre membership function and target-prediction functions are learned simultaneously. Rather than being chosen to minimize an unsupervised quantity such as within-cluster-sum-of-squares, cadres are selected to maximize the effectiveness of the target-prediction process. Only a small subset of features are used for the the cadre-assignment and target-prediction processes -- their functions are sparse in the input space. To achieve sparseness, we impose elastic net priors\cite{ElasticNet} on the supervised cadre model's parameters. The cadre membership function and each cadre's target-prediction function are allowed to use a different set of features.

%A supervised cadre model is more interpretable than nonlinear techniques such as kernel support vector machines\cite{svms} or deep neural networks\cite{deeplearning}. It is suitable for learning problems in which model predictions are meant to augment, rather than replace, human decisions and insight. The learning goal goes beyond generalization: the domain expert should be able to understand how a machine learning model makes its predictions. The proposed methods are also effective on problems for which there is insufficient data for complicated nonlinear models to perform well. 

%\textbf{ASN note:} We say this (the preceding sentence), but is it justified? In the benchmarking section, on the two smallest datasets (Airfoil and Boston), the kernel SVR outperforms the linear SVR and is about as good as the cadre model. And we don't consider $p\gg n$ regime problems.

In this paper, we develop the mathematical formalism of the supervised cadre model and its learning problem, focusing on the case where the learning task is regression. We assess the quality of a cadre structure in terms of generalization, interpretability, and subpopulation robustness. Experimental benchmarks show that the SCM has generalization that is competitive with linear, piecewise-linear, and nonlinear regression methods. The cadre models identify stable subpopulations with sparse membership and prediction functions. 

To demonstrate interpretability, we provide a cheminformatics case study in which the SCM is applied to the the problem of predicting the glass transition temperature of polymers. The cadres found represent different subpopulations of molecules based on their connectivity, polarizability, and distribution of negatively charged surface area. These features correlate with specific molecular structure characteristics, which can provide design advice for creating new polymers with desired properties\cite{design}. The cadre method identifies small subsets of features that are used uniquely by each cadres and shared features that are predictive across all cadres. 

An implementation of the SCM is available on GitHub\footnote{https://github.com/newalexander/supervised-cadres}.

\subsection{Motivating Example}
We consider a simple two-dimensional example to demonstrate the usefulness of a modeling procedure that can jointly discover subpopulations and predict the target. It is inspired by a glass temperature transition\cite{Katritzky} analysis found in Section \ref{cheminformatics}, but the data are simulated. In this problem, the cadres identify groups of molecules that obey different underlying physics. The SCM can solve this problem very well, but existing methods that combine clustering and regression can fail to learn a good prediction rule. 
%{\bf use more informative variables than x1 and x2 in plots
%Need to clean this up, put the exact things ploted in the figure caption}

%predicting glass transition temperature for polymers, where the polymers in each cadre run on different underlying physics.

\begin{figure} 
    \centering
  \subfloat[Test data, colored by connectivity subpopulation\label{toy_true}]{%
       \includegraphics[width=0.45\linewidth]{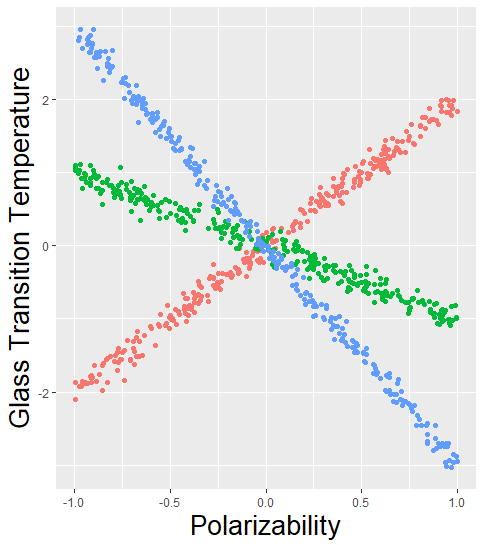}}
    \hfill
  \subfloat[Test data, colored by connectivity subpopulation\label{toy_true_hist}]{%
        \includegraphics[width=0.45\linewidth]{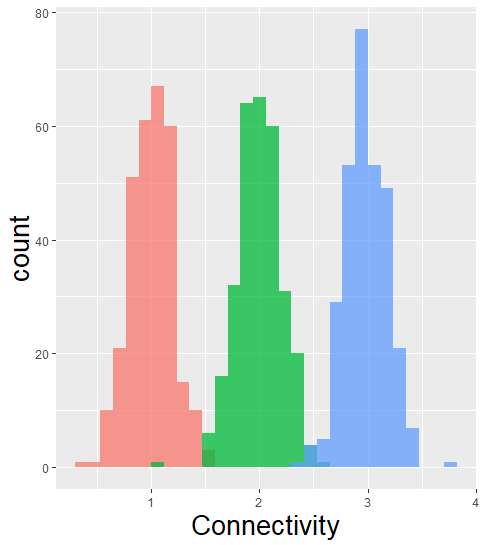}}
    \\
  \subfloat[Test data, colored by $k$-means cluster\label{toy_km}]{%
        \includegraphics[width=0.45\linewidth]{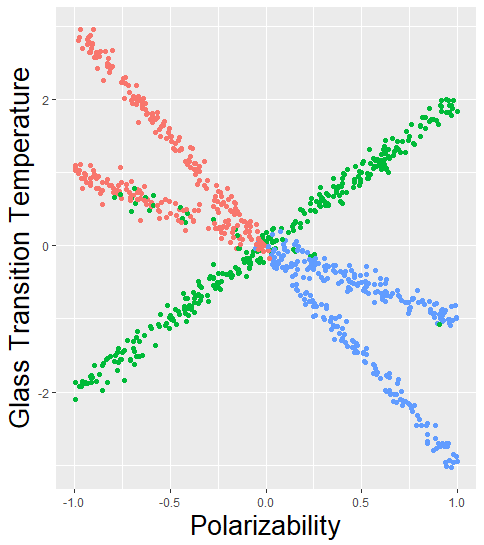}}
    \hfill
  \subfloat[Test data, colored by cadre\label{toy_cadre}]{%
        \includegraphics[width=0.45\linewidth]{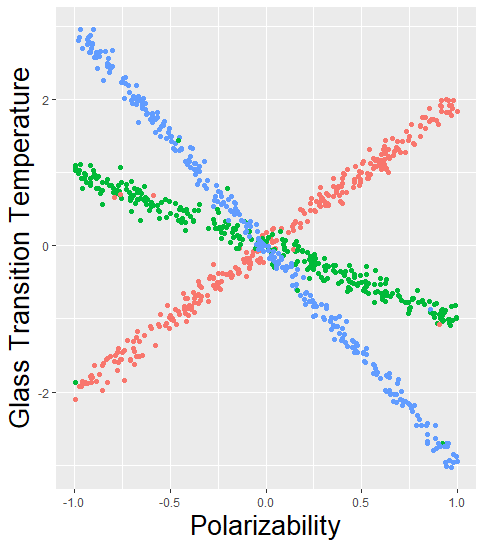}}
      
  \caption{(a) and (b) show the test data and its connectivity-based subpopulations. In (c), $k$-means learns the wrong subpopulations for the red and blue parts. $K$-means erroneously clusters jointly over connectivity and polarizability; however, the clustering mechanism that generated the data depends only on connectivity. In (d) the SCM learns the correct subpopulations because it performs feature selection while discovering cadres.}
  \label{toyPlots} 
\end{figure}

We seek to the predict the glass transition temperature of a polymer using two characteristics: its branching connectivity and electrical polarizability. Polarizability can inform understanding of a polymer's response to external fields and also relates to the level of induced electrical attraction between polymer strands; connectivity characterizes the level of backbone branching of functional groups in a polymer. For illustration purposes, imagine that, for molecules with low levels of branching and steric bulkiness, electrical polarizability is positively correlated with glass transition temperature. This means that polymer strands are able to associate with each other more strongly and remain associated at higher temperatures. For molecules with medium or high branching connectivity, the polarizability is negatively associated with glass transition temperatures, with different rates. The polymers can be grouped by their branching connectivity, as shown in Fig. \ref{toy_true} and \ref{toy_true_hist}. In Fig. \ref{toy_km} and \ref{toy_cadre}, a standard clustering method such as $K$-means\cite{lloyd} cannot discover the true subpopulations, but the SCM can.

% \begin{figure}[!ht]
% \subfloat[Test data, colored by connectivity group\label{toy_true}]{\includegraphics[width=0.25\textwidth]{toyTrue.png}}
% \subfloat[Test data, colored by connectivity group\label{toy_true_hist}]{\includegraphics[width=0.25\textwidth]{toyTrueHist.png}}\\
% \subfloat[Test data, colored by $k$-means cluster\label{toy_km}]{\includegraphics[width=0.25\textwidth]{toyKm.png}}
% \subfloat[Test data, colored by cadre\label{toy_cadre}]{\includegraphics[width=0.25\textwidth]{toyCadre.png}}
% \caption{(a) and (b) show the . In (c), $k$-means learns the wrong grouping for the red and blue parts. In (d) the supervised cadre model learns the correct grouping.}
% \label{toyPlots}
% \end{figure}

%width=1.75in

\section{Related Work}

Prior methods for combining a clustering task and a supervised learning task exist. For regression problems, there is the clusterwise linear regression model\cite{clusReg}. For classification problems, there is the clustered SVM\cite{clusSVM} and mixture of linear SVM\cite{mixSVM} models. The clusterwise linear regression and clustered SVM models learn hard partitions of the space, whereas we learn a soft partition. The idea of the mixture of linear SVMs is to approximate a nonlinear SVM with a large number of linear models. Because we want interpretable models, we are interested in problems in which a small number of carefully chosen linear models are sufficient for good prediction. In these prior models, there is no feature selection performed during the clustering and prediction processes.

The input space is probabilistically partitioned, and each partition is assigned a simple supervised learner in the hierarchical mixture of experts (HME)\cite{jordanHME}. The SCM can be viewed as an HME with only one set of gates and experts and a special gating function. The standard HME gating function cannot discover subpopulations, unlike ours.

Clustering has been combined with feature selection and weighting to yield the sparse $K$-means method\cite{sparse_cluster} and weighted fuzzy $c$-means method\cite{fuzzy}, respectively. Both are intended for problems that lack a usable target feature, and sparse $K$-means focuses on those in the $p\gg n$ regime. We restrict our attention to supervised learning problems so that the subpopulations are defined relative to the behavior of a target feature.

Recent papers \cite{Lipton}\cite{DoshiVelez} have proposed different ways to quantify a model's interpretability, including simulatability, decomposability, and amicability to post-hoc interpretation. Our model is simulatable because a human can easily replicate a model calculation. It is decomposable because all of its parameters have intuitive purposes. It also admits post-hoc interpretations because visualizations of feature distributions can be grouped by cadre for greater insight in a dataset's structure. Additionally, the SCM is interpretable because its constituent functions are sparse\cite{lasso}.

%Finally, cadre models can be regarded as an interpretable extension to SVM. The radial basis function (RBF) SVM makes predictions by taking a linear combination of many functions, each centered around a support vector. The cadre model uses relatively few RBF to define cadre membership. SVM uses constant weights on each RBF. In cadres, each RBF is weighted by a sparse linear SVM.  As in relevance vector machines \cite{RVM}, only relevant features are used to define the cadres membership function.  Cadre models use a deeper structure than SVM to maintain generalization capabilities while gaining interpretability.

\section{Model Development}

\subsection{Formalism}

Let the input be $x \in \mathbb{R}^P$, and let the target be $y \in \mathcal{Y}$. We focus on regression problems where $\mathcal{Y} = \mathbb{R}$, but cadre models can be applied to classification problems where $\mathcal{Y} = \{1,\hdots,K\}$ if the loss function is changed. 
%Let $F_P = \{1,\hdots,P\}$ be the set of feature indices. 
We make two assumptions. The first is that there exist subsets $X^1,\hdots,X^M$ of $\mathbb{R}^P$ such that the behavior of $y$ at points within each $X^m$ is more simply modeled than it is on the whole input space. We call these subsets cadres. The second is that some features affect only $x$'s cadre membership, and some features, given $x$'s cadre membership, affect only the predicted value of $y$. %We do not know which features are used for cadre-determination and which features are used for target-prediction.

%We select index sets $F_C, F_T \subseteq F_P$, with $P_C = |F_C|$ and $P_T = |F_T|$. If $p \in F_C$, then the feature $x_p$ is used to determine what cadre an observation $x$ belongs to. If $p \in P_T$, then the feature $x_p$ is used to predict the target. So we refer to $P_C$ as the cadre-assignment features, and we refer to $F_T$ as the target-prediction features. For the rest of this paper, we let $F_C = F_T = F$. That is, we make no a priori assumptions about whether a given feature is used for cadre-assignment or target-prediction. %This distinction is decided during the learning process.

We define a prediction function $f:\mathbb{R}^P \to \mathbb{R}$ that may be decomposed as
$$f(x) = \sum_m g_m(x) e_m(x)$$
where $g_m(x) = p(x \in X^m)$ and $e_m(x) = \mathbb{E}[y |x, x \in X^m]$. From these and $f$'s definition, we can show that $f(x) = \mathbb{E}[y | x]$. We parameterize each $g_m$ and $e_m$ as
$$g_m(x) = \frac{e^{-\gamma ||x - c^m||_d^2}}{\sum_{m'}e^{-\gamma ||x - c^{m'}||^2_d}}\,\,\,\,\text{and}\,\,\,\,e_m(x) = \left(w^m\right)^T x + w_0^m.$$
Here: $||x||_d = \left(\sum_p |d_p| (x_p)^2\right)^{1/2}$  is a seminorm; $d$ is a feature-selection parameter used for cadre assignment; each $c^m$ is the center of the $m$th cadre, each pair $w^m, w^0_m$ characterizes the regression hyperplane for cadre $m$; and $\gamma > 0$ is a hyperparameter that controls the sharpness of the cadre-assignment process. Thus, the cadre membership of $x$ is a multinoulli random variable with probabilities $\{g_1(x),\hdots,g_M(x)\}$ characterized by weighted automatic relevance determination (ARD)\cite{Murphy} kernels $e^{-\gamma||x-c^m||_d^2}$. We can also interpret $g_m(x)$ as the softmax\cite{Murphy} of the set of weighted inverse-distances $\{\gamma||x-c^1||_d^{-2},\hdots,\gamma||x-c^M||_d^{-2}\}$. We let $m^*(x) = \arg\max_m g_m(x)$ be the cadre that observation $x$ is most likely to belong to.

%\textbf{ASN Note: We could include a schematic showing that $f$ is sort of a shallow neural net if we have the room.}

If we let $C$ and $W$ be matrices with columns $c^1,\hdots,c^M$ and $w^1,\hdots,w^M$, our model is fully specified by the the parameters $C \in \mathbb{R}^{P_C \times M}$, $d \in \mathbb{R}^{P_C}$, $W \in \mathbb{R}^{P_T \times M}$, $w^0 \in \mathbb{R}^M$, and the hyperparameter $\gamma>0$. We group a model's parameters as $\Theta = \{C, d, W, w^0\}$. All the parameters have interpretations, ensuring model decomposability\cite{Lipton}. Each $c^m$ is the center of the $m$th cadre. The coefficient $d_p$ indicates how important the $p$th feature is for the cadre-assignment process. Each cadre has a regression hyperplane characterized by $w^m$ and $w^0_m$.

\begin{figure}[!htp]
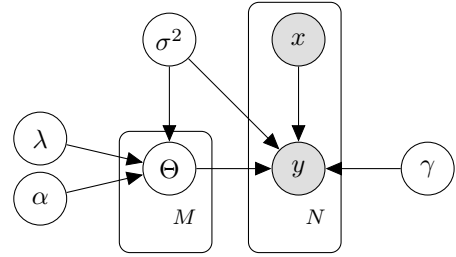

	\centering
	\tikz{
	\node[obs] (y) {$y$} ;
	\node[obs, above=of y] (x) {$x$};

	\node[latent, left=of y] (theta) {$\Theta$};
			
	\node[latent, right=of y] (gamma) {$\gamma$};
	\node[latent, above=of theta] (sigma) {$\sigma^2$};
	
	\node[latent, left=of theta, yshift=-0.4cm] (alpha) {$\alpha$};
	\node[latent, left=of theta, yshift=0.4cm] (lambda) {$\lambda$};
	
	\plate[inner sep=0.25cm, xshift=-0.05cm, yshift=-0.12cm] {plate1} {(theta)} {$M$};
	\plate[inner sep=0.25cm, xshift=-0.05cm, yshift=-0.12cm] {plate2} {(y) (x)} {$N$};

	\edge {lambda, alpha, sigma} {theta};
	\edge {theta, sigma, x, gamma} {y};
	}
	\caption{Graphical representation of supervised clustering model}\label{gr}
\end{figure}

We give a graphical representation of this model in Fig. \ref{gr}. The hyperparameters $\lambda$ and $\alpha$ and the noise variance $\sigma^2$ are described in Section \ref{implementation}.

\subsection{Learning a Cadre Model}\label{implementation}

For regression problems, we use the standard $y | x \sim \mathcal{N}(f(x), \sigma^2)$. With a different loss model, an SCM can be specialized to other supervised learning tasks. For example, hinge-loss\cite{Murphy} could be used in classification. Models are learned with Bayesian point estimation. Let $X$ be the set of training observations, let $Y$ be the set of training target values, and let $N$ be number of training observations. We learn the supervised cadre model by solving the negative log-posterior minimization problem:
\begin{eqnarray*}
\{\hat{\Theta}, \hat{\sigma}^2\} &=& \arg\min\{ -\log p(\Theta, \sigma^2 | X, Y)\}\\ 
&=& \arg\min\{ -\log(p(Y|\Theta,\sigma^2,X)p(\Theta|\sigma^2)p(\sigma^2))\}
\end{eqnarray*}
Expanding the term $-\log p(Y|\Theta,\sigma^2,X)$ will generate loss terms proportional to $\ell(x,y) = (f(x) - y)^2$. When $g_m(x)$ is close to 0 or 1, the gradient of this loss term is poorly conditioned. A tractable upper bound is $\ell(x,y) \leq \sum_m g_m(x)(e_m(x) - y)^2$. Its validity relies on the fact that $f(x)$ is a convex combination of $\{e_1(x),\hdots,e_M(x)\}$, and it is tight in the limiting case that $g_m(x) \in \{0,1\}$ for all $x$.

The prior distribution for $\Theta$ is factored as $p(\Theta | \sigma^2) = p(d|\sigma^2) p(W|\sigma^2)$. We impose elastic net priors\cite{ElasticNet} on $W$ and $d$ with hyperparameters $\lambda = \{\lambda_d, \lambda_W\}$ and $\alpha = \{\alpha_d, \alpha_W\}$:
$$p(d | \sigma^2) \propto \exp\left(-\frac{1}{2\sigma^2}R(d; \lambda_d, \alpha_d)\right)$$
$$p(W | \sigma^2) \propto \exp\left(-\frac{1}{2\sigma^2}R(W; \lambda_W, \alpha_W)\right),$$
where $R(\cdot; \lambda_i, \alpha_i) = \lambda_i(\alpha_i||\cdot||_1 + (1-\alpha_i)||\cdot||_2^2)$ is the elastic net regularization functional. Note that, for $W$, the norms are applied entrywise. The elastic net lets us find a compromise between the stability of an $\ell^2$ penalty and the sparseness of an $\ell^1$ penalty. Following the conventions of Bayesian linear regression\cite{Murphy}, we impose an uninformative improper prior on $\sigma^2$: $p(\sigma^2) = 1 / \sigma^2$. 

The final augmented loss functional, minimized with respect to $\Theta=\{d,C,W,w^0\}$ and $\sigma^2$
is
\begin{eqnarray*}
\mathcal{L}(\Theta, \sigma^2) &=& \frac{1}{2\sigma^2} \sum_{n=1}^N\sum_m g_m(x^n)(y_n - e_m(x^n))^2\\
&+& (1+N)\log \sigma^2\\
&+& \frac{\lambda_d}{2\sigma^2}(\alpha_d||d||_1 + (1-\alpha_d)||d||_2^2)\\
&+& \frac{\lambda_W}{2\sigma^2}(\alpha_W||W||_1 + (1-\alpha_W)||W||_2^2).
\end{eqnarray*}

Our augmented loss functional is nonsmooth and nonconvex, which makes it difficult to minimize with conventional optimization algorithms. We use the TensorFlow\cite{tensorflow} module for Python, which uses automatic differentiation for fast stochastic first-order optimization. Within TensorFlow, we use Adam\cite{Adam}, an adaptive stepsize gradient decent method. The stochastic aspect of our optimization process makes our solver less susceptible to local minima: a point in parameter-space that is a bad local minimizer for one minibatch of the training data is not guaranteed to be a local minimizer for other minibatches of the training data\cite{bottou}.

The learning problem is fully specified by the cadre-assignment sharpness hyperparameter $\gamma$, the elastic net mixing hyperparameters $\alpha$, the regularization strength hyperparameters $\lambda$, and the number of cadres $M$. %These may be selected via some combination of validation, domain knowledge, and user preference. 

\subsection{Model Assessment}\label{assessment}

We evaluate the overall quality of a supervised cadre method in three ways: generalization, cadre quality, and interpretability. Generalization is how well the model works on new data. It can be measured by evaluating an accuracy metric such as mean squared error on a holdout set. A cadre is good if it accurately captures a subpopulation\cite{stability}. We develop two metrics of interpretability and examine a case study in depth. These metrics are applied to real datasets in Section \ref{empirical}.

\subsubsection{Cadre Quality}  We use average best match\cite{Hopcroft} across bootstrapped samples as our metric of cadre quality.
To properly gauge cadre goodness, the nonconvexity of the cadre model requires specially chosen parameter initializations. We begin by taking a bootstrap sample $\mathcal{D}_0$ of the data and learning a cadre model $\{\Theta_0, \sigma^2_0\}$ from this sample. We then apply Algorithm \ref{boot} to learn $B$ more models. For each of the $B+1$ models, we find the cadre assignment of every observation in the original dataset. A cadre learned in bootstrap sample 0 is good if it also learned in many of the other bootstrap samples. So define $C_m^b$ to be the index set
$$C^b_m = \{n \in \{1,\hdots,N\} : \arg\max g_{m'}(x^n) = m\}.$$
Then the match between two index sets $C^b_m$ and $C^{b'}_{m'}$ is
$$\text{match}(C_m^b, C_{m'}^{b'}) = \min\left\{\frac{|C_m^b \cap C_{m'}^{b'}|}{|C_m^b|}, \frac{|C_m^b \cap C_{m'}^{b'}|}{|C_{m'}^{b'}|}\right\}.$$
Two sets $C^b_m$ and $C^{b'}_{m'}$ have a match score close to 1 when they share many elements in common and are approximately the same size. The match score decreases as either of those assumptions ceases to hold. Given a set of models from bootstrapped samples, we find the average best match\cite{Hopcroft} $ABM(m,M)$ for cadre $m$ in a model with $M$ total cadres. Average best match is given by
$$ABM(m,M) = \frac{1}{B}\sum_{b>0}\max_{m'}\{ \text{match}(C^0_m, C^b_{m'})\}.$$ 

If a cadre has a high average best match, the model has identified a stable subpopulation. Given a set of bootstrap samples, the average best match assigns a number to every cadre in model 0. If we are comparing models with different total numbers of cadres, we can assign the entire model an average best match: $ABM(M) = \frac{1}{M}\sum_m ABM(m,M)$.

\begin{algorithm}[ht]
\caption{Bootstrap Algorithm for Cadre Quality}\label{boot}
\begin{algorithmic}
\REQUIRE{Dataset $\mathcal{D}=\{(x^n,y_n)\}_{n=1}^N$, number of bootstrap trials $B$, number of cadres $M$, starting values $\{\Theta_0,\sigma^2_0\}$}
\ENSURE{A list $\{m^b\}_{b=1}^B$ of assigned cadres}%, a list $\{LL_b\}$ of log-likelihoods}
\FOR{$n=1,\hdots,N$} \STATE{$m^0_n \leftarrow \arg\max_m g_m(x^n; \Theta_0)$} \ENDFOR
\FOR{$b=1,\hdots,B$} \STATE{Let $\mathcal{D}_b$ be a bootstrap sample of $\mathcal{D}$\\
Learn a cadre model $\{\Theta_b, \sigma^2_b\}$ from $\mathcal{D}_b$, using $\{\Theta_0,\sigma^2_0\}$ as starting values} \ENDFOR
\FOR{$n=1,\hdots,N$} \STATE{$m^b_n \leftarrow \arg\max_m g_m(x^n; \Theta_b)$} \ENDFOR
\end{algorithmic}
\end{algorithm}

%\begin{algorithm2e}
%\Kw{Dataset $\mathcal{D}=\{(x^n,y_n)\}_{n=1}^N$, number of bootstrap trials $B$, number of cadres $M$, starting values $\{\Theta_0,\sigma^2_0\}$}
%\KW{A list $\{m^b\}_{b=0}^B$ of assigned cadres}%, a list $\{LL_b\}$ of log-likelihoods}
%\For{$n = 1,\hdots,N$}{
%$m^0_n \leftarrow \arg\max_m g_m(x^n; \Theta_0)$}
%\For{$b = 1,\hdots,B$}{
%Let $\mathcal{D}_b$ be a bootstrap sample of $\mathcal{D}$\\
%Learn a cadre model $\{\Theta_b, \sigma^2_b\}$ from $\mathcal{D}_b$, using $\{\Theta_0,\sigma^2_0\}$ as starting values\\
%%Let $LL_b$ be the log-likelihood of $\mathcal{D}$ from $\{\Theta^*, \sigma^2_*\}$\\
%\For{$n = 1,\hdots,N$}{
%$m^b_n \leftarrow \arg\max_m g_m(x^n; \Theta_b)$}
%}
%\caption{Bootstrap Algorithm for Testing Stability}\label{boot}
%\end{algorithm2e}

\subsubsection{Interpretability} We separately measure the interpretability of an SCM's cadre-assignment and target-prediction processes. For cadre-assignment, we consider how easily a domain expert can understand and validate the discovered subpopulations -- i.e., post-hoc interpretability. This is easiest when the model is sparse\cite{Lipton}. Sparsity is summarized with the density rate $DR=||d||_0/P$, where $||d||_0$ is the number of nonzero entries in $d$ and $P$ is $d$'s dimension.

For target-prediction, we introduce a statistic $\tau$ that measures how similar the set of learned linear models are to one another. When $\tau$ is small, then the linear models that each cadre has are very similar to each other. When $\tau$ is large, the linear models are very different. So $\tau$ provides high-level information about how each cadre's regression weights vary from each other. We quantify this similarity by
$$\tau = \frac{1}{P}\sum_{p=1}^P\text{StdDev}(\{w^1_p,\hdots,w^M_p\}).$$
So $\tau$ is the standard deviation of every model's weight for the $p$th feature, averaged over all features.

\section{Empirical Evaluation}\label{empirical}

In our empirical evaluation, first we present an in-depth case study on a cheminformatics problem. Then we present accuracy and cadre quality benchmarks on a variety of datasets. In all numerical results, we begin by centering and standardizing both the input and target data. The elastic net mixing parameter $\alpha$ was fixed at $\alpha_d = 0.95$ and $\alpha_W = 0.05$. All other hyperparameters, including those for competitor methods, were set by cross-validation over the training set.

\subsection{Cheminformatics Case Study}\label{cheminformatics}

We demonstrate how the supervised cadre model may be applied to a regression problem. The task is to predict a polymer's glass transition temperature (Tg), given various measurements of its characteristics\cite{Katritzky}\cite{WuThesis}. 

Polymers transition from a brittle state to a viscous state if they are heated past their glass transition temperature. The glass transition temperature thus characterizes pure polymers, polymer blends, and copolymers. It can also be informative of a polymer's miscibility. The relationship between the target and measurable features is known to be nonlinear. In addition, although a single glass transition temperature is measured, in truth, there are a range of temperatures over which a polymer undergoes its glass transition. Thus, predictions of transition temperature need not be perfectly accurate. In addition, previous work has suggested that polymers may possess a natural cadre structure\cite{WuThesis}.

In our polymer dataset, there were 262 different polymers. We took an 80\%-20\% train-test split. The dataset had 119 features initially. Because of feature correlation, we followed \cite{WuThesis} and applied a correlation filter (threshold 0.7) to the training data. After this, 28 features remained. 

\begin{figure} 
    \centering
\subfloat[Test set MSE of SCM (purple line) and piecewise-linear models.\label{chem_m}]{%
        \includegraphics[width=0.37\linewidth]{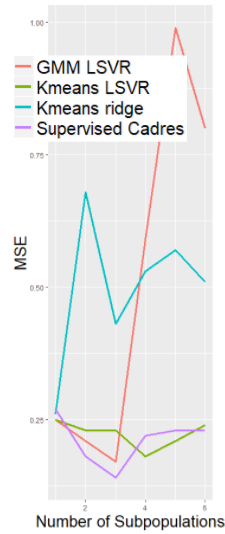}}
    \hfill
      \subfloat[Predicted vs. true Tg values for full dataset, colored by cadre membership.\label{tgYvsY}]{%
       \includegraphics[width=0.53\linewidth]{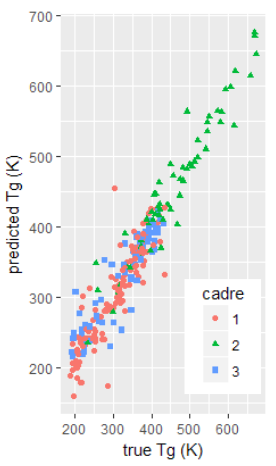}}
\caption{Results from initial cheminformatics analysis. In (a), as number of subpopulations increases, MSE hits an optimal minimum and then plateaus or increases. In (b), the cadre structure splits observations based on their Tg value.}
\end{figure}

We compared the SCM's predictive ability to that of other methods and determined the most natural number of cadres via cross-validation. These results are presented in Table \ref{table:chem} and Fig. \ref{chem_m}. The models compared were the SCM, the $K$-means linear SVR, the $K$-means regression tree, the Gaussian Mixture linear SVR, the random forest, and the RBF SVR. In the $K$-means models, a $K$-means clustering was learned on the training set. Then, for each of the clusters, a separate supervised learning model was trained on the observations in each cluster. The Gaussian mixture linear SVR was defined similarly. A full covariance matrix was learned. The SCM outperforms the other models on this task.
%We see that, at $M=3$ cadres, the supervised cadre model does best. %We show how test set MSE varies as the number of subpopulations changes in Fig. \ref{chem_m}.

\begin{table}[!htp]
%\resizebox{0.5\textwidth}{!}{
\begin{tabular}{l|l|l}
Model & MSE & $M^*$\\\hline
Supervised cadres & \textbf{0.14} & 3\\
$K$-means ridge regression & 0.26 & 1\\
$K$-means linear SVR & 0.18 & 4\\
$K$-means regression tree & 0.24 & 2\\
Gaussian Mixture linear SVR & 0.17 & 3\\
Random forest & 0.21 & 1\\
RBF SVR & 0.22 & 1\\
\end{tabular}
%}
\caption{Generalization error and model complexity on Tg dataset. }\label{table:chem}
\end{table}

Now we discuss the subpopulations found by the SCM. Fig. \ref{tgYvsY} shows the true Tg values versus their predicted values. Cadres 1 and 3 (respectively, the red circles and blue squares) contain polymers with low Tg. Cadre 2 (the green triangles) primarily contains polymers with high Tg. We provide the mean and standard deviation Tg of each cadre in Table \ref{table:chem2}.

% \begin{figure}[!htp]
% \begin{center}
% \includegraphics[height=2in]{trueTg_predTg.png}
% \caption{Predicted vs. true Tg values, colored by cadre membership. The cadre structures splits observations based on their Tg value.}\label{tgYvsY}
% \end{center}
% \end{figure}

\begin{table}[!htp]
%\resizebox{0.5\textwidth}{!}{
\begin{tabular}{l|l|l|l}
Cadre & \# Polymers & Mean Tg (K)& Std.Dev Tg\\\hline
1 & 121 & 300 & 68\\
2 & 62 & 456 & 106\\
3 & 79 & 330 & 72
\end{tabular}
%}
\caption{Cadre Characteristics. Std.Dev Tg is the standard deviation of the distribution of Tg scores in each cadre.}\label{table:chem2}
\end{table}

\begin{figure}[!htp]
\begin{center}
\includegraphics[height=2.4in]{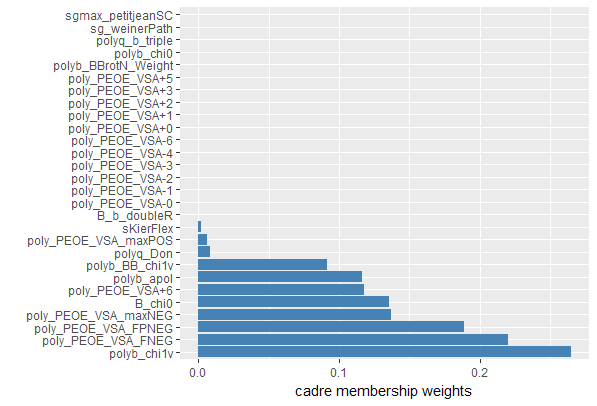}
\caption{Distribution of cadre assignment weights $d_p$. Few features were used for cadre-assignment.}\label{memWei}
\end{center}
\end{figure}

We can also consider which features are important for determining cadre membership and how these features vary between cadres. Fig. \ref{memWei} shows the distribution of cadre membership weights $d_p$. The optimal $d_p$ was very sparse: only 11 of the 28 features were used in the cadre-assignment process. So the cadre assignment process was simple.

\begin{figure}[!htp]
\subfloat[]{\includegraphics[width=1.5in]{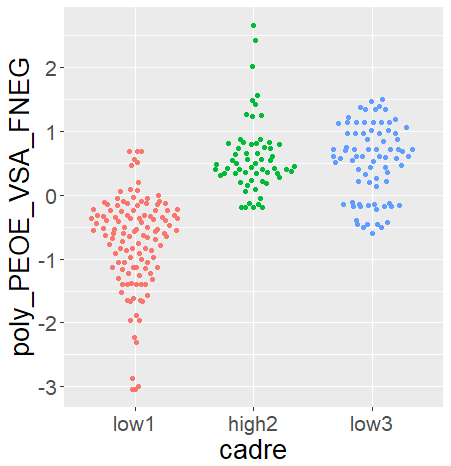}}
\subfloat[]{\includegraphics[width=1.5in]{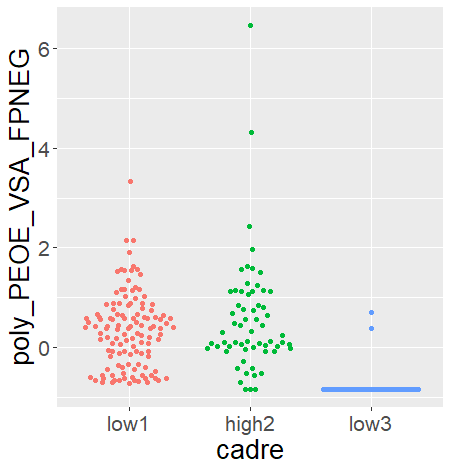}}\\
\subfloat[]{\includegraphics[width=1.5in]{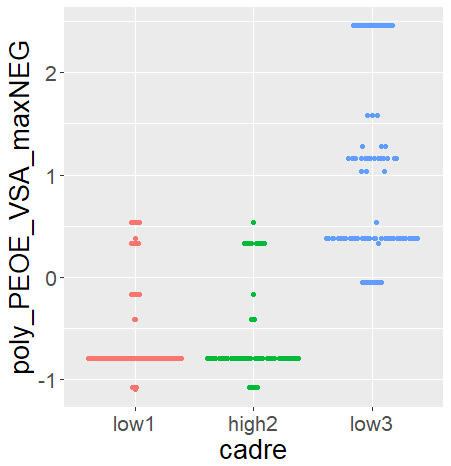}}
\subfloat[]{\includegraphics[width=1.5in]{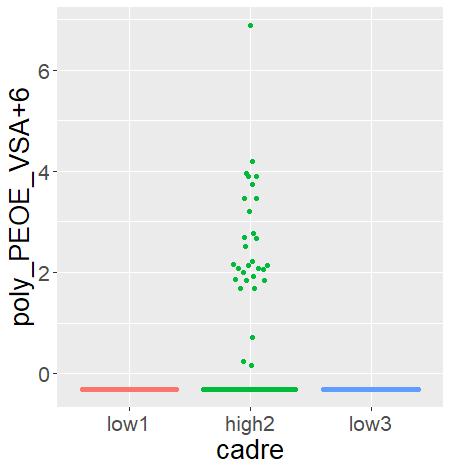}}\\
\subfloat[]{\includegraphics[width=1.5in]{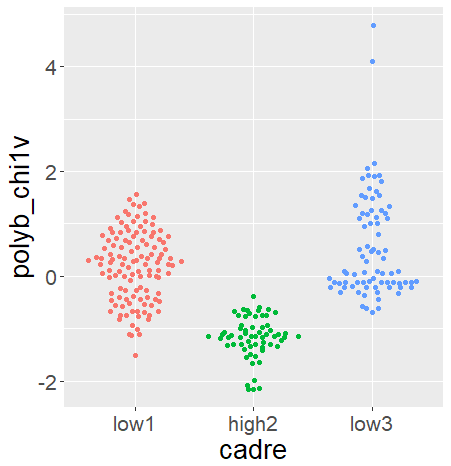}}
\subfloat[]{\includegraphics[width=1.5in]{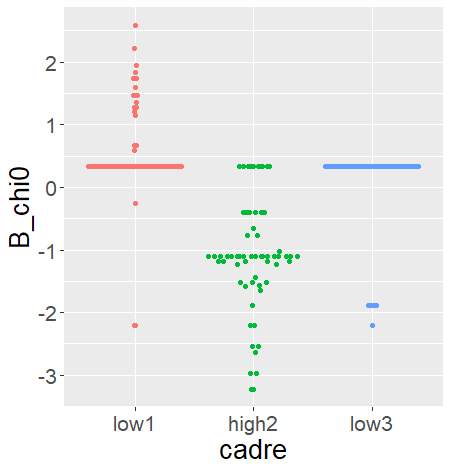}}\\
\subfloat[]{\includegraphics[width=1.5in]{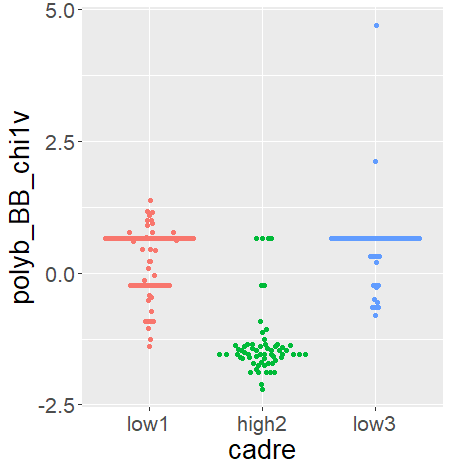}}
\subfloat[]{\includegraphics[width=1.5in]{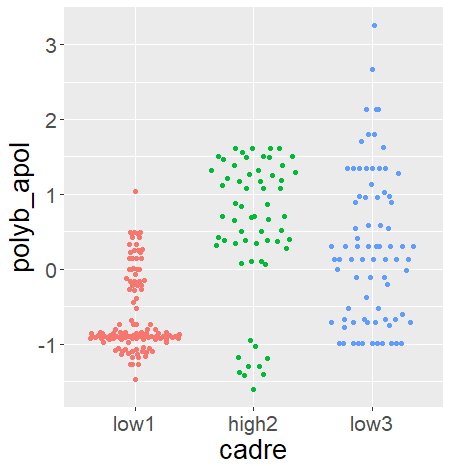}}
\caption{Distribution of features for each cadre, colored by cadre. Partial charge [(a)-(d)] features mostly distinguish between the two low-Tg cadres. Connectivity [(e)-(g))] and polarizability (h) features distinguish between the two low-Tg cadres and between the low-Tg and high-Tg cadres.}
\label{chemfeatures}
\end{figure}

Looking at Fig. \ref{memWei}, the features important for determining cadre membership may be grouped into three categories: partial charge distribution, connectivity, and atomic polarizability. Four features related to partial charge surface distributions were important for cadre membership: poly\_PEOE\_VSA\_FNEG, poly\_PEOE\_VSA\_FPNEG, poly\_PEOE\_VSA\_maxNEG, and poly\_PEOE\_VSA+6. Three of them are used to distinguish between the two low-Tg cadres. Three features related to molecular bulk and backbone branching were important for cadre membership: polyb\_chi1v, B\_chi0, and polyb\_BB\_chi1v. The final important feature was polyb\_apol, a descriptor associated with induced electrostatic attraction between polymer chains. This examination of features shows how post-hoc interpretation helps to distinguish between the two low-Tg cadres and between the low-Tg cadres and the high-Tg cadres. In Fig. \ref{chemfeatures}, the feature distributions grouped by cadre reflect distinct subpopulations.

\begin{figure}[!htp]
\includegraphics[width=3.3in]{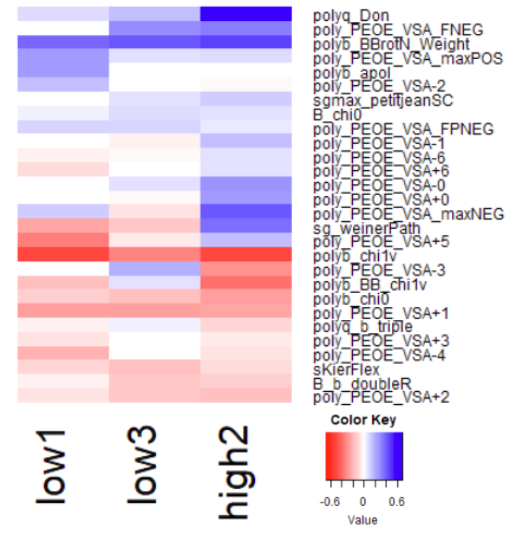}
\caption{Distribution of cadre regression weights $w^m_p$. The different cadres learned different linear models.
}\label{predWeights}
\end{figure}

Now we consider the regression models for each cadre. The weights $w^m_p$ are shown in Fig. \ref{predWeights}. These weights have a $\tau$ score of 0.12: the cadres use distinct models but share some features. Some features, such as polyb\_BBrotN\_Weight (the weighted effect of rotatable bonds in the polymer repeat unit) and poly\_PEOE\_VSA+1 (related to surface charge), are used similarly in all three cadres. Some features are used differently between cadres. This includes sg\_weinerPath, a connectivity index, which is positively associated with Tg in the high-Tg cadre but negatively associated with Tg in the low-Tg cadres. Some features are not used by every cadre, such as poly\_PEOE\_VSA-3, which is positively associated with Tg in cadre 3, negatively associated with Tg in cadre2, and not used in cadre 1. These sparsity patterns help further characterize the differences between the two low-Tg cadres and the high-Tg cadre. Some features, such as polyb\_PEOE\_VSA+6, polyb\_PEOE\_VSA\_FPNEG, and B\_chi0, are very important for determining cadre membership but have small regression weights. We compare the polymers in cadres 1 and 3 to characterize their differences. Cadre 1 contains many elastic acrylates, but cadre 3 is primarily alkenes. The brittle, high-Tg carbonates are all placed into cadre 2. Within each cadre, the underlying patterns of physical factor controlling the Tg of each polymer are different and are represented by different but overlapping sets of descriptors. 

The descriptor patterns prominent in each cadre represent sets of latent variables that describe how the Tg of a polymer within that cadre would respond to specific structural changes. This ability to separate polymers with different sensitivities to structural alterations into cadres enables domain experts to design materials with desirable Tg properties using models with expanded and identifiable domains of applicability\cite{design}.

%\textbf{Can we fit a concluding thought here? Or should it go in the conclusion?}

\subsection{Accuracy Benchmarks}\label{accuracy}

We want to empirically evaluate the SCM. Thus, we apply it to a variety of datasets taken from the UCI\cite{UCI} repository. The datasets were Airfoil Self Noise (Airfoil)\cite{AirfoilSelfNoise}, AirQuality\cite{AirQuality}, Boston Housing (Boston)\cite{Boston}, Concrete Compressive Strength (Concrete)\cite{Concrete}, Facebook Comment Volume (Facebook)\cite{Facebook}, SkillCraft1 Master Table (GameSkill)\cite{GameSkill}, and Parkinson's Telemonitoring (Parkinson)\cite{Park}. We give the number of features and observations in Table \ref{table:sizes}. Note that the Parkinson dataset contains two targets; we learn separate models for each.

% we can add ||d||_0 if we want

\begin{table}[!htp]
\resizebox{\columnwidth}{!}{
\begin{tabular}{l|l|l|l|l|l|l}
Dataset & $N$ & $P$ & $M^*$ & $ABM$ & $DR$ & $\tau$\\\hline
Airfoil & 1503 & 6 & 4 & 0.64 & 0.66  & 0.19\\
AirQuality & 9358 & 15 & 2 & 1.00 & 0.31 & 0.05\\
Boston & 506 & 13 & 5 & 0.81 & 0.50 & 0.08\\
Concrete & 1030 & 8 & 5 & 0.83 & 0.66 & 0.10\\
Facebook & 40949 & 39 & 4 & 0.79 & 0.43 & 0.04\\
GameSkill & 3395 & 20 & 4 & 0.62 & 0.17 & 0.03\\
Parkinson1 & 5875 & 16 & 4 & 0.82 & 0.34 & 0.07\\
Parkinson2 & 5875 & 16 & 4 & 0.79 & 0.35 & 0.08\\
\end{tabular}}
\caption{For benchmark datasets, number of observations $N$, number of features $P$, optimal number of cadres $M^*$, model-level average best match $ABM$, model density rate $DR$, and average weight standard deviation $\tau$. The $DR$ and $\tau$ values are averaged over different dataset instances.}\label{table:sizes}
\end{table}

Our experimental design is as follows: We generated 20 different 75\%-25\% train-test splits for each dataset. For each split, we trained a model and recorded its mean squared error (MSE) when evaluated on the test set. The models we used were: a cadre model, a linear SVR (LSVR), an RBF SVR (KSVR), a $K$-means linear SVR (KM+LSVR), and a Gaussian Mixture Linear SVR (GMM+LSVR). By $K$-means linear SVR, we mean that we learned a $K$-means clustering on the training data and then trained a linear SVR on the observations in each cluster. The GMM+LSVR model is defined similarly. The Gaussian Mixture Model learned a full covariance matrix for each mixture component. Throughout, model selection was performed with cross-validation on the training set.

%The models we used were: a cadre model, a linear SVR model (lsvr), a Gaussian kernel SVR model (ksvr), a ridge regression model (ridge), a $k$-means linear SVR model (kmLsvr), and a $k$-means ridge regression model (kmRidge). By $k$-means linear SVR (ridge regression), we mean that we learned a $k$-means clustering on the training data, and then trained a linear SVR (ridge regression) on the observations in each cluster. This is thus a hard partition of the input-space.

For every dataset and every model, we calculated the mean and standard deviation of the test set MSE scores. These are shown in Table \ref{table:many_errors}. In the AirQuality and GameSkill datasets, the SCM had a significantly lower ($p<0.05$) mean test error than all other methods. In the Airfoil, Concrete, Parkinson1, and Parkinson2 datasets, the SCM had a significantly lower ($p<0.05$) mean test error than all methods save KSVR. In the Facebook dataset, the SCM had the lowest mean test error of all methods, but the differences were not significant. This shows that the SCM is competitive with powerful nonlinear regression methods and has performance that often exceeds naive piecewise-linear regression methods.

% \begin{table*}[h]
% \centering
% \caption{Error statistics for empirical evaluation. SCM is the supervised cadre model, GMM+LSVR is the Gaussian mixture model combined with linear SVR, KM+LSVR is $k$-means combined with linear SVR, KSVR is RBF SVR. The first number is the mean test set MSE; the second number is the standard deviation of test set MSE values.}
% \label{table:many_errors}
% \begin{tabular}{llllll}
% Dataset    & SCM             & GMM+LSVR        & KM+LSVR          & KSVR            & LSVR            \\
% Airfoil    & 0.2502 (0.0444) & 0.4114 (0.0413) & 0.4300 (0.0420) & 0.2207 (0.0221) & 0.4857 (0.0318) \\
% AirQuality & 0.0001 (0.0001) & 0.0007 (0.0001) & 0.0030 (0.0002) & 0.0020 (0.0002) & 0.0035 (0.0003) \\
% Boston     & 0.2009 (0.0873) & 0.2240 (0.0657) & 0.2199 (0.0660) & 0.2025 (0.0640) & 0.3283 (0.0904) \\
% Concrete   & 0.1858 (0.0348) & 0.2696 (0.0423) & 0.2417 (0.0261) & 0.1640 (0.0284) & 0.4170 (0.0501) \\
% GameSkill  & 0.4174 (0.0202) & 0.4421 (0.0151) & 0.4318 (0.0205) & 0.4327 (0.0216) & 0.4929 (0.0404) \\
% Park1      & 0.8082 (0.0232) & 0.8663 (0.0341) & 0.8626 (0.0346) & 0.7233 (0.0317) & 0.9308 (0.0483) \\
% Park2      & 0.7926 (0.0338) & 0.8985 (0.0437) & 0.8923 (0.0414) & 0.7429 (0.0386) & 0.9327 (0.0450)
% \end{tabular}
% \end{table*}

\begin{table}[ht]
\centering
\label{table:many_errors}
\resizebox{0.5\textwidth}{!}{
\begin{tabular}{l|l|l|l|l|l}
Dataset    & SCM         & GMM+LSVR    & KM+LSVR      & KSVR        & LSVR        \\\hline
Airfoil    & 0.25 (0.04) & 0.41 (0.04) & 0.43 (0.04) & \textbf{0.22 (0.02)} & 0.48 (0.03) \\
AirQuality & \textbf{1e-4 (1e-4)} & 7e-4 (1e-4) & 3e-3 (2e-4) & 2e-3 (2e-4) & 4e-3 (3e-4) \\
Boston     & \textbf{0.20 (0.08)} & 0.22 (0.06) & 0.21 (0.06) & \textbf{0.20 (0.06)} & 0.32 (0.09) \\
Concrete   & 0.18 (0.03) & 0.26 (0.04) & 0.24 (0.02) & \textbf{0.16 (0.02)} & 0.41 (0.05) \\
Facebook   & \textbf{0.68 (0.18)} & 0.78 (0.20) & 0.76 (0.20) & 0.75 (0.20) & 0.78 (0.20) \\
GameSkill  & \textbf{0.41 (0.02)} & 0.44 (0.01) & 0.43 (0.02) & 0.43 (0.02) & 0.49 (0.04) \\
Parkinson1 & 0.80 (0.02) & 0.86 (0.03) & 0.86 (0.03) & \textbf{0.72 (0.03)} & 0.93 (0.04) \\
Parkinson2 & 0.79 (0.03) & 0.89 (0.04) & 0.89 (0.04) & \textbf{0.74 (0.03)} & 0.93 (0.04)
\end{tabular}}
\caption{Mean and standard deviation of test set MSE values for Section \ref{accuracy}. Bolded entries are the best for a particular dataset. The SCM performs significantly better ($p<0.05$) than all of the methods except KSVR, but SCM is interpretable and KSVR is a black box.
}
\end{table}

% \begin{figure}[!ht]
% \begin{center}
% \includegraphics[width=3.5in]{all_accs_2.PNG}
% \caption{Test set mean MSEs. The error bars are the standard deviations of the mean MSE distributions. The cadre model has orange lines. It is always the best or second best model. Not all differences are statistically significant. %Note that AirQuality uses a different $y$-axis scale than the other datasets.
% }\label{allErrors}
% \end{center}
% \end{figure}

%\FloatBarrier

\subsection{Cadre Goodness and Interpretability Evaluation}

We check that the SCM is capable of learning robust and interpretable cadres. This section's results used the optimal number of cadres $M^*$ found in Section \ref{accuracy} via cross-validation. For each of the datasets detailed in Section \ref{accuracy}, we generate 20 different bootstrap samples. For each split, we learn an SCM and evaluate our cadre goodness and interpretability metrics on that model; the results are in Table \ref{table:sizes}. The average best matches (ABM) show that the SCM can learn stable cadre structures. The average $||d||_0$ (DR) shows that most models had quite sparse cadre-assignment mechanisms. The average variance $\tau$ measures how distinct the regression weights were: the Airfoil model had very distinct weights, and AirQuality, Facebook and GameSkill models had similar weights.

The SCM is more interpretable than the competitors considered here. Unlike the SCM, Linear SVR, $K$-means, and gaussian mixtures are not sparse over the feature space. RBF SVR is nonlinear and nonparametric, so it is the least interpretable of any model considered here.

\section{Conclusion}
We have developed a new framework for discovering informative subpopulations (cadres)  while solving a predictive task. Here, we have focused on regression problems and linear prediction functions, but our notion of a cadre is applicable to learning problems in general. In an in-depth cheminformatics case study, we showed that the regression cadre model learns post-hoc interpretable groupings of polymers that each have accurate, informative models. The method identified cadres of polymers that respond differently to structural perturbations, thus providing design insight for targeting or avoiding specific transition temperature ranges. 

We assessed the quality of a cadre structure in terms of generalization, interpretability, and subpopulation robustness. On benchmark datasets, SCM performs well with respect to these metrics. For generalization, it outperforms linear and piecewise-linear models and is competitive with nonlinear ones. Due to its sparse linear aspects, it is more interpretable than non-sparse piecewise-linear and fully nonlinear models.

In the future, cadre models can be customized to other subpopulation definitions, base functions, and learning tasks by changing the gating,  emission and loss functions. We are currently applying cadre analysis to other classes of learning problems including classification and risk analysis. The key contribution is that SCM can discover informative subpopulations and thus enable a new type of precision analysis. This can augment expert knowledge in fields as diverse as drug discovery, fault detection, and health risk analysis.

%On benchmark datasets, our model performs well with respect to these metrics. For generalization, it beats linear and piecewise-linear models and is competitive with nonlinear ones. Because of its sparse linear aspects, it is more interpretable than non-sparse piecewise-linear and fully nonlinear models.

%-- the notion of a cadre. Cadres are characterized by a sparse probabilistic partition of input space, and each cadre is assigned its own linear model. . We have presented three key metrics for evaluating a cadre model: generalization, stability, and interpretability. On benchmark datasets, our model performs well with respect to these metrics. For generalization, it beats linear and piecewise-linear models and is competitive with nonlinear ones. Because of its sparse linear aspects, it is more interpretable than non-sparse piecewise-linear and fully nonlinear models. In an in-depth cheminformatics case study, we have shown that the supervised cadre mode learns chemically interpretable groupings of polymers. These groupings allow domain experts to augment their understanding of the factors that affect a polymer's glass transition temperature. We can also apply the ideas of cadre analysis to discover informative subpopulations in classification and risk-analysis problems.

\section*{Acknowledgment}

Thanks to NSF Grant \#1331023 and Dr. Ke Wu for support of this project. This work is supported by IBM Research AI through the AI Horizons Network and by the Center for Biotechnology \& Interdisciplinary Studies at Rensselaer. 

% \IEEEtriggeratref{8}
\bibliographystyle{IEEEtran}
\bibliography{sdmBib}
\end{document}